\documentclass[runningheads]{llncs}
\usepackage[T1]{fontenc}
\usepackage{graphicx}
 \usepackage{nicematrix}
 \usepackage{multirow}
\usepackage{array,multirow,graphicx}
\usepackage{tabularx}
\usepackage{caption}
\usepackage{booktabs}
\usepackage{hyperref}

\begin{document}

\def\methodName{SANGRIA}
\title{\methodName{}: Surgical Video Scene Graph Optimization for Surgical Workflow Prediction}
\titlerunning{SANGRIA}
%
%
\author{}
\institute{}

\author{Çağhan Köksal \inst{1, 2}$^*$ \and
Ghazal Ghazaei\inst{1}$^*$ \and Felix Holm \inst{1,2} \and  Azade Farshad \inst{2,3} \and Nassir Navab \inst{2,3} }

\authorrunning{C. Koksal, G. Ghazaei, F. Holm, A. Farshad, N. Navab. }
\def\thefootnote{*}\footnotetext{Equal contribution. Corresponding author email: caghankoksal@gmail.com}\def\thefootnote{\arabic{footnote}}
\def\thefootnote{**}\footnotetext{This work was conducted and fully financed by the Corporate Research and Technology department of Carl Zeiss AG.}\def\thefootnote{\arabic{footnote}}
\institute{Carl Zeiss AG, Germany \and
Technische Universität München, Germany \and
Munich Center for Machine Learning, Germany}


%
%
\maketitle              

%

%
%
%
\begin{abstract}
Graph-based holistic scene representations facilitate surgical workflow understanding and have recently demonstrated significant success. However, this task is often hindered by the limited availability of densely annotated surgical scene data. In this work, we introduce an end-to-end framework for the generation and optimization of surgical scene graphs on a downstream task. Our approach leverages the flexibility of graph-based spectral clustering and the generalization capability of foundation models to generate unsupervised scene graphs with learnable properties. We reinforce the initial spatial graph with sparse temporal connections using local matches between consecutive frames to predict temporally consistent clusters across a temporal neighborhood. By jointly optimizing the spatiotemporal relations and node features of the dynamic scene graph with the downstream task of phase segmentation, we address the costly and annotation-burdensome task of semantic scene comprehension and scene graph generation in surgical videos using only weak surgical phase labels.
Further, by incorporating effective intermediate scene representation disentanglement steps within the pipeline, our solution outperforms the SOTA on the CATARACTS dataset by $8\%$ accuracy and $10\%$ F1 score in surgical workflow recognition.



\end{abstract}

\section{Introduction}
\label{sec:intro}

Surgical videos capture pivotal moments of surgery, providing a valuable source of information that can facilitate better insights into the quality of surgery. 
Automated analysis of these videos can significantly enhance surgical procedures via online or offline feedback. Surgical workflow prediction has been a focal point of numerous studies, highlighting its critical role in enhancing surgical precision and efficiency through video analysis \cite{twinanda2016endonet,shah2023glsformer,bertasius2021space,sharma2023rendezvous,holm2023dynamic,murali2023encoding} 
Recently, methods based on scene graph representations overperformed non-graph methods thanks to its holistic scene understanding capabilities\cite{holm2023dynamic,sharma2023surgical,murali2022latent,farshad2023scenegenie,murali2023encoding}. A primary barrier to developing such technologies is the lack of dense annotations.
Moreover, surgical video annotation is not only inherently burdensome but requires specialized annotation platforms and expert annotators, which can be prohibitively expensive for specialized surgeries. Given the inherent interdependence of surgical scene understanding and workflow prediction during surgical procedures, it is imperative to address these two tasks simultaneously. In this regard, we introduce a novel approach of \textbf{S}urgic\textbf{A}l Sce\textbf{N}e \textbf{GR}aph Opt\textbf{I}miz\textbf{A}tion (SANGRIA) that tackles both problems using solely surgical phase labels. 

\begin{figure*}[t]
  \centering
   \includegraphics[width=\linewidth]{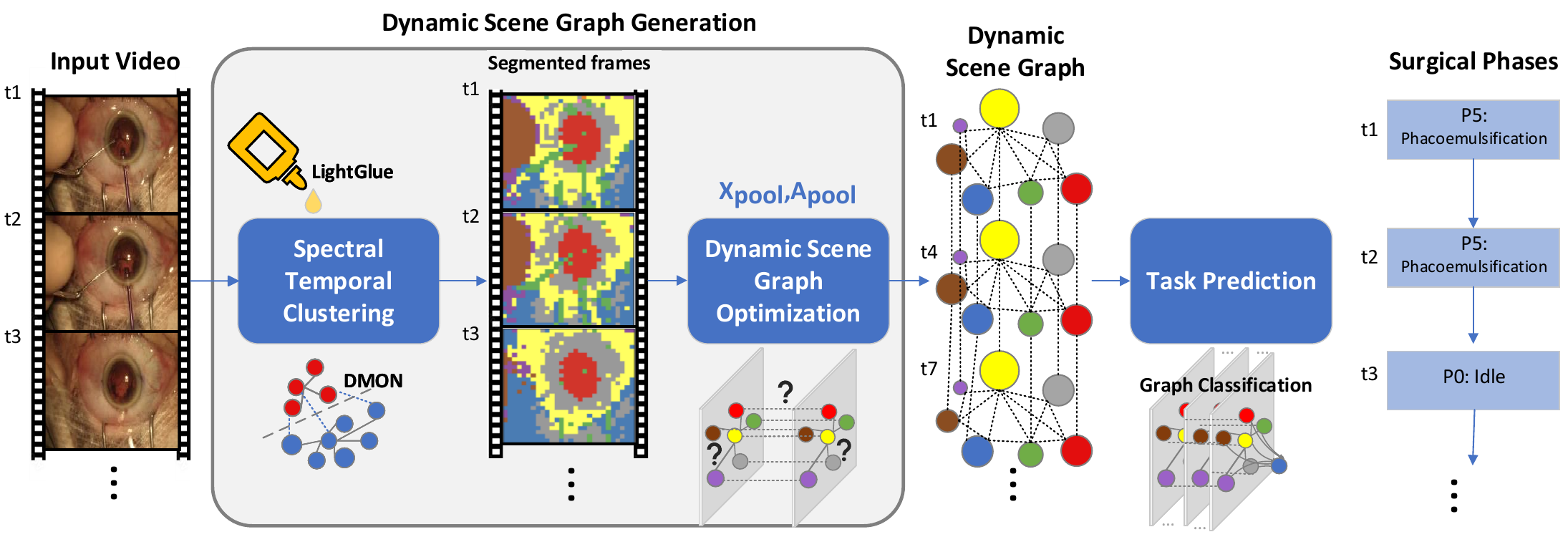}

    \caption{Our end-to-end surgical scene graph generation and workflow prediction pipeline, \methodName{}, comprising: 1) \textbf{Spectral Temporal Clustering} converts input frames into a dynamic patchified graph leveraging graph partitioning and local feature matching to produce an initial dynamic scene graph. 2) \textbf{DSG Optimization} optimizes the edge weights between clusters for the end task. 3) \textbf{Task prediction} a GCN-based architecture to predict surgical phases.}
   \label{fig:teaser}
\end{figure*}

Prior works on unsupervised scene and video segmentation take advantage of optical flow \cite{francesco2020object} or similar to CutLER \cite{wang2023cut,wang2023videocutler} use variations of NormalizedCut \cite{shi2000normalized} and self-training \cite{wang2023tokencut} to find salient objects in the scene. For surgical scene understanding, similarly, optical flow or shape priors have been used \cite{zhao2020learning,sestini2023fun}. Shape priors lack adaptability and flexibility to new tools or setups and struggle when visually similar tools are present. While optical flow is more generalizable, it introduces noise from anatomical movement or fluid during surgery. 
The main shortcoming with solutions such as \cite{wang2023cut,wang2023videocutler} is the burdensome process of mask generation followed by rounds of self-training, which would be needed on every new dataset. 
The advent of foundation models \cite{caron2021emerging,oquab2023dinov2,kirillov2023segment} has  opened new opportunities for scene understanding with minimal annotations. In the context of surgical scenes, this impact is still less pronounced, considering the significant domain gap between common computer vision use cases and medical applications, requiring further fine-tuning.
Previous research demonstrated that (dynamic) scene graph representations provide a more holistic and interpretable representation of surgical procedures for surgical workflow recognition \cite{holm2023dynamic,murali2023encoding}. 
To mitigate the annotation scarcity in surgical videos and address both scene understanding and surgical workflow recognition simultaneously, we propose a novel data-agnostic pipeline that integrates the benefits of semantic scene graphs and graph-based unsupervised semantic scene segmentation techniques. Our approach comprises two main components: 

\textit{First}, a task- and data-agnostic spectral and temporal clustering module leveraging DMON layer \cite{tsitsulin2023graph} for unsupervised clustering of scene components reinforced with LightGlue \cite{lindenberger2023lightglue} for sparse and lightweight local feature matching among neighboring frames. Acting as a zero-shot semantic scene segmenter, this module ensures local and temporal consistency while optimizing scene representation for downstream tasks.
Our solution builds upon insights from recent works \cite{aflalo2023deepcut,bianchi2020spectral}, formulating semantic segmentation as a graph partitioning task. While these models indicate promising performance on still images, they lack temporal consistency among neighboring frames when applied to video sequences. 
To overcome this challenge, we leverage LightGlue \cite{lindenberger2023lightglue} into our dynamic graph construction, which predicts an assignment between points based on their pairwise similarity and unary matchability. This leads to sparse dynamic links among close frames and significantly reduced computations. 

\textit{Second}, a dynamic scene graph (DSG) generation module optimizing graph relations based on the downstream task. The end-to-end pipeline is jointly optimized for the specific downstream task, which is surgical phase segmentation in this work.
As the predicted semantic segmentation maps and the generated scene graphs lack object class identities, we propose a simple yet efficient prototype matching strategy via which only a few prototypes are exploited to retrieve class identities. 
In this work, we introduce SANGRIA, a novel graph-based pipeline harnessing the holisticness and flexibility of graph-based learning to overcome the challenge of annotation scarcity in surgical video analysis. 

Our contributions are as follows: 
1) Unsupervised semantic scene segmentation and dynamic scene graph generation leveraging a spectral and temporal clustering module equipped with lightweight correspondence matching. 2) Demonstrating the importance of understanding scene components and their spatiotemporal relations in contributing to the downstream task. By disambiguating video representations in an end-to-end optimization setup, SANGRIA achieves state-of-the-art performance on surgical phase segmentation on the CATARACTS \cite{al2019cataracts} dataset. 3) Offering a few-shot prototype matching mechanism enabling further refinement of the predictions and demonstrating promising performance in annotation-efficient scene graph generation.

\section{Methodology}
\label{sec:method}
We tackle the problem of semantic scene understanding and scene graph generation with an end-to-end graph-based pipeline. 
Formulating semantic segmentation as a graph partitioning task, we patchify input images and establish sparse temporal links with the neighboring frame patches via correspondence matching, constructing a dynamic graph (the patch-based graph hereafter).
We then perform a spectral, temporal clustering of the patch-based graph to generate a dynamic semantic scene graph. This DSG is augmented with a dynamic relation prediction module to be further refined for the downstream task of surgical phase segmentation.
Finally, a prototype matching mechanism is developed for a final refinement, few-shot segmentation, and scene graph generation evaluations.

\subsection{Dynamic Scene Graph Generation}


For an input image $I$, DINO\cite{caron2021emerging} key features ${f}$ are obtained by partitioning ${I}$ into $n$ patches and passing them to DINO. The adjacency matrix ${A}$ is then generated by patchwise dot product as follows:

\begin{equation} \label{eq:similarity}
    {\mathcal{A}}_{ij} = 
    \begin{cases} 
    {f}_i \cdot {f}_j & \text{if } {f}_i \cdot {f}_j > 0  \\
    0 & \text{otherwise} 
    \end{cases}
\end{equation}
Next, we threshold the values in ${\mathcal{A}>\tau}$ and connect highly similar nodes together, generating a static patch graph representation $\mathcal{G}_t = (\mathcal{V}_t,\mathcal{A}_t)$ for a given frame t.
An extension of graph clustering to sequences of frames without considering the temporal inter-dependencies leads to inconsistent clusters among frames with close proximity. A na\"ive solution could be an expansion of adjacency matrix calculation across the third dimension of time to find spatiotemporal similarities across patches. This leads to high computational costs $\mathcal{O}(w n^2 d$) specifically with the increasing length of the temporal window for $n$ number of patches, patch feature of length $d$, and $w$ time steps. As temporal relations require a coarser level of attention compared to spatial dependencies~\cite{carreira2017quo}, we suggest a sparse dynamic linking mechanism between patches along the time dimension. 

In this work, we leverage correspondence matching to find prominent features within frames and match those efficiently. We incorporate LightGlue \cite{lindenberger2023lightglue}, a distilled deep neural network powered with self- and cross-attention, into our patch graph construction setup. It is designed explicitly for low-latency problems and sparse inputs by predicting matches from two sets of local features.
Next,  for a clip with $w$ frames, we construct a dynamic patch-based graph, $\mathcal{G}_{t_i\xrightarrow{}t_{i + w}} = (\mathcal{V},\mathcal{E})$ with node set $\mathcal{V}$, edge set $\mathcal{E}$, node features $X\in \mathbb{R}^{w \times n \times d}$. Spatial edges, $\mathcal{E}_{t_i}$ are established using pairwise correlation similarity between those features (Equation~\ref{eq:similarity}), while for temporal edges $\mathcal{E}_{t_i \xrightarrow{}t_{i + 1}}$, LightGlue correspondences between frames within a temporal sequence of $w$ time stamps are exploited. Dynamic graph edges can be represented as follows: $\mathcal{E} =  \bigcup_{1 \leq t \leq w}  \mathcal{E}_{t_i}  + \mathcal{E}_{t_i \xrightarrow{}t_{i + 1}} $

We further reinforce the graph nodes with temporal and spatial encodings to accentuate the dynamic relations between objects in the scene. Temporal encodings capture the temporal order of object interactions and actions in a video sequence, while spatial encodings capture objects' relative positions and orientations in a scene.
For temporal encoding, we incorporate the location of each frame along the temporal window by adding a temporal feature vector to the node feature matrix $X$. For spatial encoding, we incorporate the position of patches within the frame by adding a spatial feature vector to feature matrix $X$.

\begin{figure}[t]
  \centering
   \includegraphics[width=\linewidth]{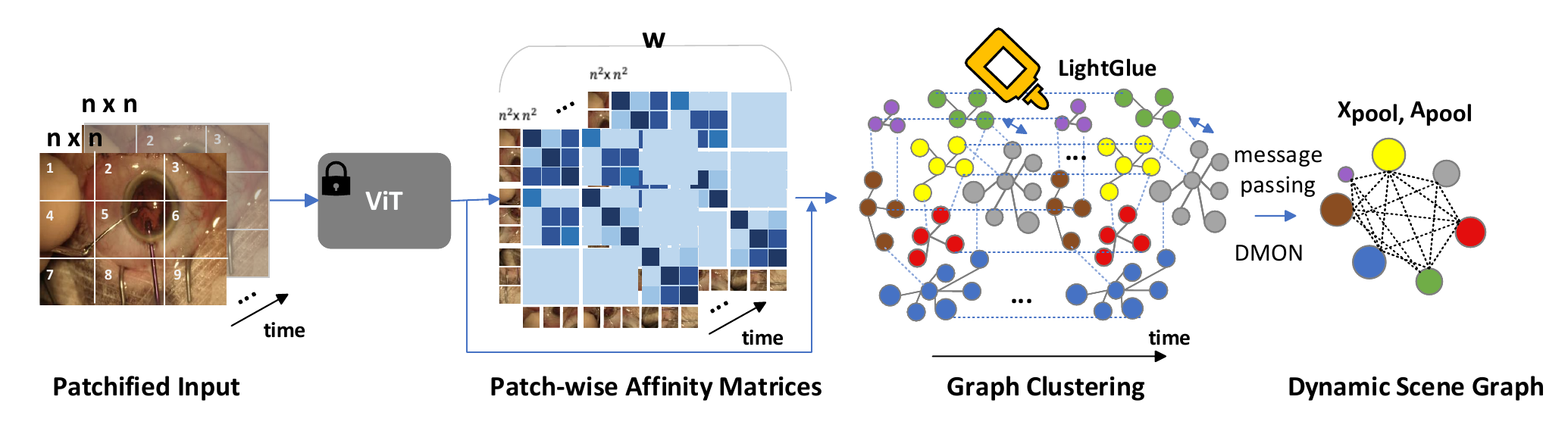}
   \caption{\textbf{DSG generation:} A patchified input image is fed into DINO to construct a patch-wise affinity matrix. The static patch-based input graphs for neighboring frames within a window of $w$ are temporally linked via sparse matches provided by LightGlue \cite{lindenberger2023lightglue}. The dynamic patch-based graph is then clustered to predict a DSG for the last frame of the window.}
   \label{fig:DSG}
\end{figure}

The graph clustering is performed by employing deep modularity networks (DMON) \cite{tsitsulin2023graph} featuring a collapse regularization objective to improve unsupervised graph clustering in real-world scenarios. The DMON module comprises multiple graph convolutional layers~\cite{kipf2016semi}, MLP layers with softmax, as well as a SeLU activation function. The input graph $X$ is fed into the DMON module to be clustered, optimizing the modularity and collapse regularization objectives.
This leads to a pooled scene graph $G^{pool}_{t_i\xrightarrow{}t_{i + w}} = (\mathcal{V}^{pool}_{t_i\xrightarrow{}t_{i + w}},\mathcal{E}^{pool}_{t_i\xrightarrow{}t_{i + w}})$ with pooled node features $X^{pool}_{t_i\xrightarrow{}t_{i + w}}\in \mathbb{R}^{K\times d}$ and pooled adjacency matrix $A^{pool}_{t_i\xrightarrow{}t_{i + w}} \in \mathbb{R}^{K \times K}$, where $K$ is the number of clusters. This DSG incorporates the summary of a window, that is, the scene components, their relations, and motion along the time. Figure~\ref{fig:DSG} illustrates the DSG generation workflow. 


\vspace{4pt}\noindent\textbf{DSG Optimization}
We establish a relaxed estimation of edge weights within the DSGs via equation $W^{pool}_{t_i \rightarrow t_{i+w}} = \sigma(\text{MLP}(X^{pool}_{t_i \rightarrow t_{i+w}}; \Theta_{\text{MLP}}))$, where $W^{pool}_{t_i\xrightarrow{}t_{i + w}}$ refers to the edge weights between the clusters of DSG, in which $w_{ij} \in [0,1] $ indicating the strength of the relations between clusters $i$, $j$. $\Theta_{\text{MLP}}$ represents the set of trainable parameters. The relaxed setup allows for flexibility in optimizing the edge weights while learning the downstream tasks and equips the DSG generation to account for the inherent uncertainty and variability present in the unsupervised graph clustering results, leading to more robust and accurate inference.


\subsection{End-to-end Pipeline}
To tackle the task of phase segmentation, we propose a multi-layer GCN~\cite{kipf2016semi,holm2023dynamic} that takes the DSG as input. The GCN consists of multiple layers, each of which enables learning increasingly complex representations of the scene graph. The output of the GCN is fed to a global sum-pooling layer aggregating features from all nodes in the graph. A fully-connected layer and a softmax function predict probabilities for each phase class.
A cross-entropy objective function $L_{CE}$ is employed to optimize the model parameters for the surgical phase segmentation task. The final objective function of the end-to-end pipeline can therefore be formulated as $L_{joint} = L_u + L_{CE}$. The first term corresponds to DMON loss, while the second term can be replaced with any other downstream task optimization loss for each specific use case. This joint optimization provides a tailored representation of the scene for the downstream task and reduces the noises introduced during unsupervised steps by taking advantage of the classification labels.

\vspace{4pt}\noindent\textbf{Prototype Matching} To assign semantic classes to DSG nodes, we create prototypes \cite{ma2013prototype} by leveraging DINO patch features and ground truth (GT) segmentation annotations. 
Exploiting only 5 annotations per class, patch features corresponding to GT segmentation masks are used as prototype patches. We use the mean of the patch features as a prototype for that object class. To predict the semantic category of a node (cluster), a pairwise cosine similarity with prototypes is calculated. Since different clusters might represent the same region, such as surgical tape, we used argmax to assign class labels to clusters.


\section{Implementation Details}
\textbf{Datasets} We experiment on 3 datasets: CATARACTS~\cite{al2019cataracts} consists of 50 cataract surgery videos of $1920 \times 1080$ pixels at $30$ fps. The dataset is split 25-5-20 for training, validation, and testing, with videos annotated on 19 surgical phases.
CaDIS dataset, a subset of CATARACTS, consists of $4670$ pixel-wise annotated images. We use Task II of CaDIS, which defines $17$ classes of objects, including surgical tools, anatomical structures, and miscellaneous.  
Cataract101 (C101) comprises 101 videos of $720 \times 540$ pixels and 25 fps performed by surgeons with various levels of expertise. The videos are annotated based on the 11 most common phases of cataract surgery and used with 45-5-50 train-validation-test splits. \\
\textbf{Training details}
Frames are resized to 224x224 to generate DINO-B embeddings. For graph generation, a frame similarity threshold of 0.9 is chosen. We trained phase segmentation models for 100 epochs using an Adam optimizer with a learning rate of 0.0001 and a batch size of 32 on a single A40 GPU.

\vspace{4pt}\noindent\textbf{Metrics} For phase segmentation, we compute the accuracy and F1 score. For semantic segmentation, we measure the mean intersection over union (mIoU) and pixel-wise accuracy (PAC).

\section{Results \& Discussion}

\textbf{Surgical Workflow Prediction} \label{res:workflow}
\autoref{tab:surgical_phase_ablation_results_dmon} presents an ablation study on window size and spatial and temporal embeddings as well as a comparative analysis of our proposed method against existing techniques on phase segmentation tasks for CATARACTS and Cataract101 datasets. We show that increasing window size together with temporal embeddings improves phase segmentation performance, while spatial embeddings have minimal impact.
Our method demonstrates superior performance in terms of accuracy and F1 score by indicating $8\%$ accuracy and $10\%$ F1 score improvement over previous graph-based phase segmentation SOTA~\cite{holm2023dynamic}. 
As there are no other benchmarks on the CATARACTS dataset, we add an additional non-graph-based solution representing the best practices in the surgical workflow. It exploits DINO as a strong spatial feature extractor followed by long-range temporal learning via TCN++\cite{li2020ms}. Our dynamic end-to-end setup outperforms the CNN baseline with $6\%$ accuracy and $4\%$ F1 score.
Our method consistently yields closely competitive results on Cataract101. This demonstrates the robustness and adaptability of our approach, providing a solid foundation for further refinement and application in diverse contexts. 
\begin{table}[tb]
   \centering
   \caption{ Comparison of DMON-based \textbf{Phase segmentation}  performance on CATARACTS \cite{al2019cataracts} and CAT101 \cite{schoeffmann2018cataract} datasets. * indicates our implementation.}
   \begin{NiceTabular}{cccccccc}
     \hline
      & Method & Graph &Spatial&Temp&WS &Accuracy & F1\\ \cline{2-8}

    \Block{14-1}<\rotate>{CATARACTS} 
    &{DINO-TCN++}* \cite{li2020ms} & Non-graph  & & &full video& 77.02 & 74.37\\
     &Holm \etal \cite{holm2023dynamic} & Static &&&1&64.34 & 50.04\\
     &Holm \etal \cite{holm2023dynamic}&Dynamic &\cmark&\cmark &30& 75.15 & 68.56 \\
     \cline{2-8}
      
      &\methodName{}  & Static   & &   &1&77.53 & 69.29\\
    &\methodName{} & Dynamic   &           &   & 4&75.50 & 67.76\\
    &\methodName{} & Dynamic  &           &   & 8&74.18 & 68.50\\
    &\methodName{}   & Dynamic &\cmark     &   &4&75.96 & 67.07\\
    &\methodName{}   & Dynamic &       &\cmark &4& {80.67} & {75.35}\\
    &\methodName{}   & Dynamic & \cmark   &\cmark  &4&81.62 & 74.62\\
    &\methodName{}  & Dynamic  & \cmark   & &8& 74.44 & 64.44\\
    &\methodName{}  & Dynamic  &   &\cmark  &8& 81.85 & 78.47\\
    &\methodName{}   & Dynamic & \cmark  &\cmark &8& 82.13 & 75.17\\
    &\methodName{}   & Dynamic & &\cmark  &16& \textbf{83.36} & \textbf{78.24}\\
    
    \hline
 
    \Block{4-1}<\rotate>{C101} 
     &ViT  \cite{dosovitskiy2020image}  &Non-graph && &1&84.56 & -\\
     &TimesFormer  \cite{bertasius2021space} & Non-graph && &8&90.76 & -\\
     &GLSFormer  \cite{shah2023glsformer} & Non-graph && &8&\textbf{92.91} & -\\
     
    &\methodName{} (Ours)  & Dynamic & & &8& 91.26  & 85.02 \\ 
    \hline
   \end{NiceTabular}
   \label{tab:surgical_phase_ablation_results_dmon}
 \end{table} 
 
\begin{figure}[t]
    \centering
    \includegraphics[width=0.94\textwidth]{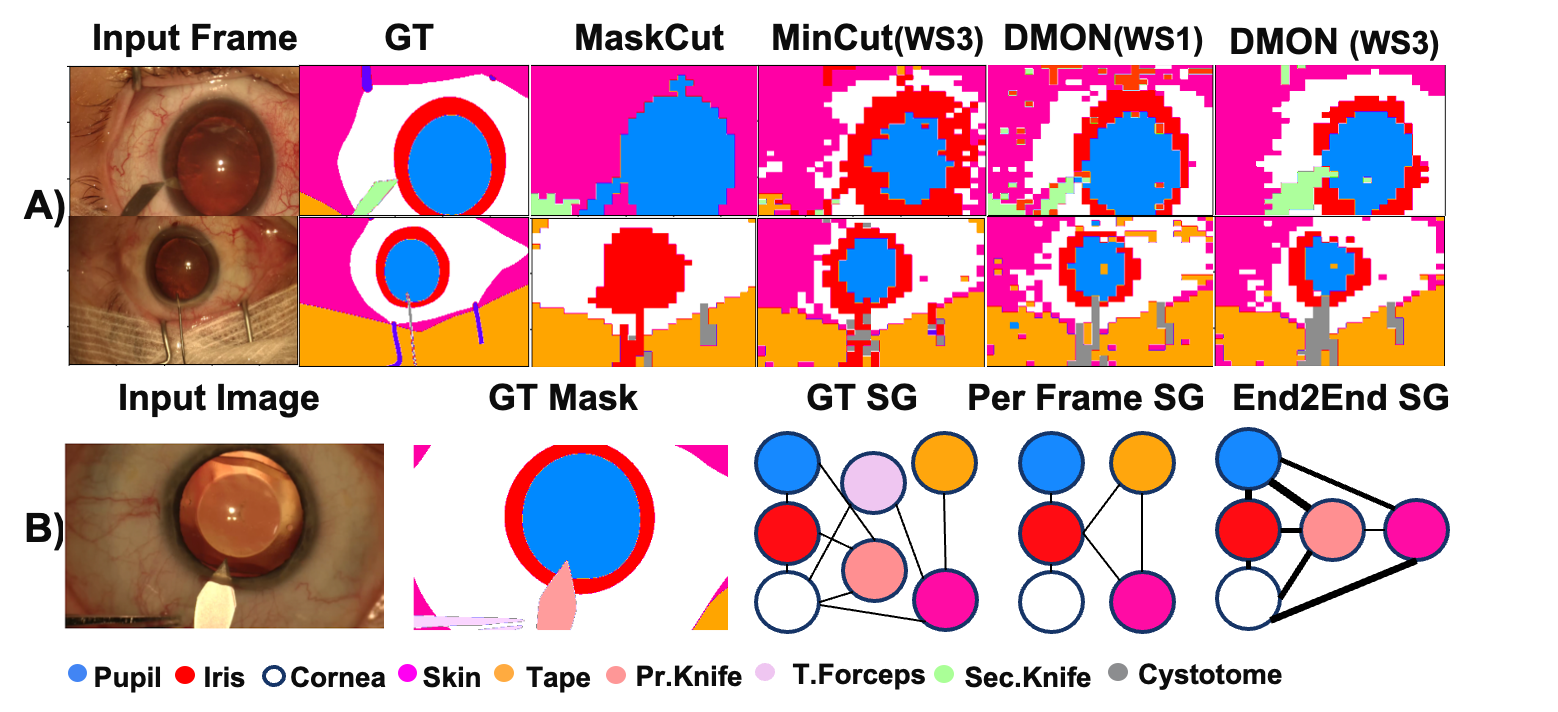}
    \caption{A) Comparison of various graph clustering setups (WS corresponds to window size). B) End-to-end optimization improves the classification of 'Primary Knife' since it plays a critical role in predicting the current phase.}
    \label{fig:qualitative}
\end{figure}

\textbf{Semantic Segmentation} \label{res:segmentation}
\autoref{tab:ablation_temporal_connections} indicates that increasing the number of temporal connections (lowering temporal similarity threshold) in fully-connected graph increases segmentation performance on both anatomy and tools.
Conversely, due to high computational costs with the increasing length of the temporal window, fully-connected graphs have limited use for end tasks since surgical phase recognition performance heavily depends on temporal information.
LightGlue correspondences effectively facilitate better temporal learning while maintaining comparable performance and considerably reducing the amount of computation.
\autoref{tab:comparison_segmentation_results_cadis_sequences} indicates the prior performance of the unsupervised segmentation performance of our proposed dynamic graph clustering algorithm to SOTA object discovery method MaskCut \cite{wang2023cut}. 
Both tables emphasize the difficulty of tool localization via mIoU$_{ins}$ while mIoU$_{ana}$ indicates higher segmentation performance for anatomical structures. In all setups, DMON consistently outperforms other solutions.
In \autoref{fig:qualitative}~A, we qualitatively evaluate MaskCut \cite{wang2023cut} and variations of our segmentation solution. Our method leverages temporal connections and predicts temporally consistent semantic segmentation maps. Unlike MaskCut, it also predicts separate masks for anatomy and surgical tools.\\

\textbf{DSG Generation} \label{res:DSG}
\autoref{fig:qualitative}~B manifests improvement of joint optimization of DSG generation with phase segmentation on surgical tool segmentation. The jointly optimized model classifies the "Primary Knife" better than the per frame optimized model. We hypothesize that guidance towards the downstream task significantly improves attention to tools with high importance in the current surgical phase. Our DSG also indicates better explainability by highlighting the importance of graph edges in the end task.

    

\noindent 
\begin{minipage}[t]{0.48\textwidth}
\vspace{0pt} 
    \centering
    \captionof{table}{Ablation study on the temporal connections.}
    \label{tab:ablation_temporal_connections}
    \resizebox{\textwidth}{!}{%
        \begin{tabular}{@{}ccccccc@{}}
    \toprule
    WS &Temp. Sim & Temp. Conn& PAC  & mIoU &  mIoU$_{ana}$ &mIoU$_{ins}$ \\
    \midrule

     1 & -        & -  & 65.34 & 33.86 & \textbf{51.44}& 9.46  \\
     3 & -        & LightGlue & 65.25 & 34.03& 50.36 & 8.68    \\
     3 & 0.9 & FC & 65.21 & 33.79 & 51.27  & 9.51  \\
     3 & 0.5 & FC & \textbf{65.37} & \textbf{34.09} & 51.50 & \textbf{9.98} \\
    
     
    \bottomrule
  \end{tabular}
    }

    
     

\end{minipage}%
\hfill 
\begin{minipage}[t]{0.48\textwidth}
\vspace{0pt} 
    \centering
    \captionof{table}{Comparison of semantic segmentation on CaDIS sequences.}
    \label{tab:comparison_segmentation_results_cadis_sequences}
    \resizebox{\textwidth}{!}{%
        \begin{tabular}{@{}ccccccc@{}}
    \toprule
    Pool Method & WS & Temp. Conn& PAC  & mIoU &  mIoU$_{ana}$ &mIoU$_{ins}$ \\
    \midrule

     Maskcut*     & 1 & - & 60.56  &  24.75 & 33.34 &6.23  \\
     MincutPool  & 1 & -  & 61.77 & 29.82 & 46.00 & 4.32\\
     DMON        & 1 & -  & \textbf{65.34} & \textbf{33.86} & \textbf{51.44} & \textbf{9.46}  \\
     \hline
     MincutPool  & 3 & LightGlue& 62.66 & 29.84 & 45.04 & 3.20  \\
     
     DMON            & 3 & LightGlue & \textbf{65.25} & \textbf{34.03}& \textbf{50.36} & \textbf{8.68}  \\


     
     
    \bottomrule
  \end{tabular}
    }

\end{minipage}

\section{Conclusion}
We introduce \methodName{}, an end-to-end graph-based solution for concurrent surgical workflow recognition, semantic scene segmentation, and dynamic scene graph generation. Our jointly optimized setup featuring sparse temporal connections and graph clustering, prioritizes the graph generation for the downstream task by disambiguating the graph and highlighting the most influential components and their connections. By focusing on downstream task-specific features,  we achieve state-of-the-art results in surgical phase segmentation on the CATARACTS dataset while generating scene explanations with minimal annotation.

\clearpage

\section{SANGRIA: Supplementary Material}

\begin{table}[h]
\centering

\label{tab:ocrnet_multi_label_classification_comparison}
\caption{Comparison of fully-supervised segmentation(OCRNET\cite{ocrnet})-based \cite{holm2023dynamic} and SANGRIA (Ours) on scene graph node classification on CaDIS sequences. While the supervised model demonstrates superior performance, our weakly-supervised approach offers promsing results on most influential objects within a surgical scene and highlight the importance of their interactions.}
\begin{tabular}{@{}c|ccc|ccc|ccc|cc@{}}
\toprule
 \multirow{2}{*}{Class} & \multicolumn{3}{c}{ Base\cite{holm2023dynamic}} & \multicolumn{3}{c}{End-to-end} \\ 
 & Precision & Recall & F1-Score & Precision & Recall & F1-Score & Support \\ 
\midrule

Pupil        & 1.00  & 1.00  & 1.00 &    1.00& 0.98 & 0.99 & 1381 \\
Tape         &  0.94 &  0.96 &  0.95    & 0.93 & 0.47 & 0.62 &  1101 \\
Hand         & 0.64  & 0.32  & 0.43  & 0.00 & 0.00 & 0.00 &  218 \\
Retractors   & 0.93 & 0.91 & 0.92 &   1.00 & 0.02 & 0.04 &  1030 \\
Iris         & 1.00  & 1.00 & 1.00   & 1.00 & 0.78 & 0.88 &  1382 \\
Skin         & 1.00 & 1.00 & 1.00   & 1.00 & 0.97 & 0.98 &  1381 \\
Cornea       & 1.00 & 1.00 & 1.00  & 1.00 & 0.99 & 0.99 &  1382 \\
Cannula      & 0.80  & 0.43 & 0.56  & 0.95 & 0.07 & 0.12 &  628 \\
Cystotome    & 0.34 & 0.55 & 0.42    &  0.06 & 0.18 & 0.09 &  33 \\
T. Forceps   & 0.77 & 0.43 & 0.55 &  1.00 & 0.00 & 0.02 &  227 \\
Pr. Knife    & 0.79 & 0.32 & 0.46    & 0.66 & 0.14 & 0.24 &  228 \\
Phaco        & 0.50  & 1.00 & 0.67    & 0.03 & 0.03 & 0.33 &  4 \\
Lens Inj.    & 0.86 & 0.45 & 0.59   & 0.98 & 0.27 & 0.42 &  223 \\
I/A          &0.19  & 0.82 & 0.31&  0.50 & 0.29 & 0.37 &  19 \\
Sec. Knife   &  0.66  & 0.30 & 0.42   & 1.00 & 0.01 & 0.02 &  187 \\
Micromanip.  &  0.14 & 0.45 & 0.22   & 0.00 & 0.00 & 0.00 &  13 \\
Cap. Forceps &  0.08 & 0.15 & 0.11   & 0.00 &0.00  & 0.00 &  20 \\
\hline
micro avg    & 0.94 & 0.87 & 0.91   & 0.96 & 0.61  & 0.75   &  9457 \\
macro avg    & 0.68 & 0.65 & 0.62   & 0.65 & 0.32  & 0.34   &  9457 \\
weighted avg & 0.94 & 0.87 & 0.89  & 0.95 & 0.61  & 0.66   &  9457 \\

\bottomrule
\end{tabular}

\end{table}

\begin{figure}[h]
    \centering
    \includegraphics[width=\linewidth]{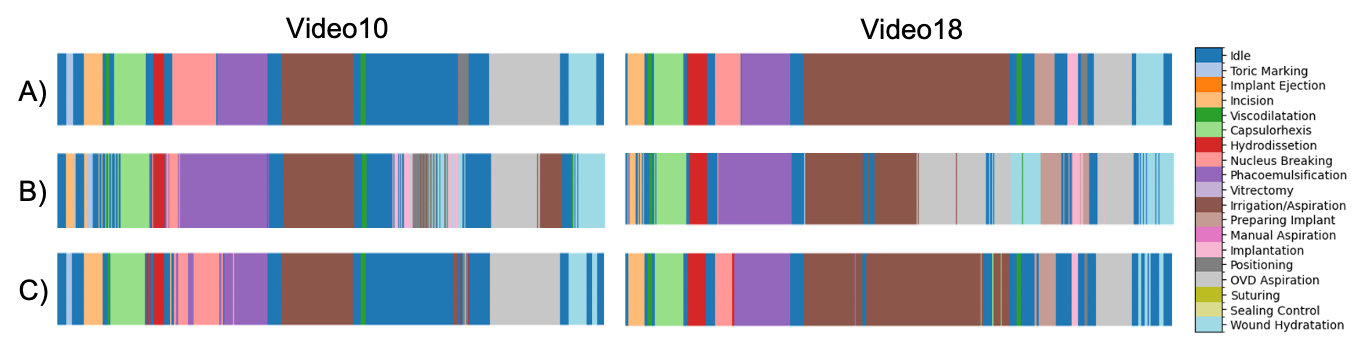}
    \caption{Comparison of phase segmentation performance on CATARACTS~\cite{al2019cataracts} test videos. A) Ground Truth phases.  B) Predictions of the best dynamic model of \cite{holm2023dynamic}. C) Best model of the SANGRIA(Ours). Our model predicts surgical phases such as Irrigation/Aspiration, OVD Aspiration and Nucleus Breaking more consistently thanks to scene representation and understanding capabilities of the SANGRIA.}
    \label{fig:qualitative_semantic_segmentation}
\end{figure}

\begin{table}[h]
\centering
\caption{Comparison of per frame optimization of MincutPool \cite{bianchi2020spectral} and DMON \cite{tsitsulin2023graph} with our end2end solution on scene graph node classification of CaDIS sequences. The end-to-end model significantly improves the node classification performance of the Lens Injector, Irrigation/Aspiration tools, which are the main components in OVD Aspiration, Irrigation/Aspiration, and Implant Injection phases. DMON consistently demonstrates better scene graph node classification performance than MincutPool.
}
\resizebox{\textwidth}{!}{%
\begin{tabular}{@{}c|ccc|ccc|ccc|ccc|cc@{}}
\toprule
 \multirow{2}{*}{Class}& \multicolumn{3}{c}{MincutPool} & \multicolumn{3}{c}{Dmon} & \multicolumn{3}{c}{End-to-end} \\ 
 & Precision & Recall & F1-Score & Precision & Recall & F1-Score & Precision & Recall & F1-Score & Support \\ 
\midrule

Pupil        & \textbf{1.00} & 0.93 & 0.97 & \textbf{1.00} & 0.96 & 0.98 & \textbf{1.00} & \textbf{0.98} & \textbf{0.99} & 1381 \\
Tape         & \textbf{0.94} & 0.54 & 0.69 & 0.93 & \textbf{0.60} & \textbf{0.73} & 0.93 & 0.47 & 0.62 &  1101 \\
Hand         & 0.91 & 0.29 & 0.44 & \textbf{0.95} & \textbf{0.38} & \textbf{0.54} & 0.00 & 0.00 & 0.00 &  218 \\
Retractors   & \textbf{1.00} &0.03  & 0.05 & 1.00 & \textbf{0.03} & \textbf{0.06} & \textbf{1.00} & 0.02 & 0.04 &  1030 \\
Iris         & \textbf{1.00} &0.71  & 0.83 & \textbf{1.00} & \textbf{0.81} & \textbf{0.90} & \textbf{1.00} & 0.78 & 0.88 &  1382 \\
Skin         & \textbf{1.00} &0.88  & 0.94 & \textbf{1.00} & 0.94 & 0.97 & \textbf{1.00} & \textbf{0.97} & \textbf{0.98} &  1381 \\
Cornea       & 1.00 &0.94  & 0.97 & 0.90 & 0.98 & 0.99 & \textbf{1.00} & \textbf{0.99} & \textbf{0.99} &  1382 \\
\hline
Cannula      & \textbf{0.96} &0.04 & 0.08 & 0.93 & 0.02 & 0.05 & 0.95 & \textbf{0.07} & \textbf{0.12} &  628 \\
Cystotome    & \textbf{0.09} &0.15 & \textbf{0.11} & 0.08 & 0.15 & \textbf{0.11} & 0.06 & \textbf{0.18} & 0.09 &  33 \\
T. Forceps   & 0.98 &0.18 & 0.30 & 1.00 & \textbf{0.30} & \textbf{0.47} & \textbf{1.00} & 0.00 & 0.02 &  227 \\
Pr. Knife    & 0.55 &0.10 & 0.17 & 0.54 & 0.11 & 0.19 & \textbf{0.66} & \textbf{0.14} & \textbf{0.24} &  228 \\
Phaco        & 0.00 &0.00 & 0.00 & 0.02 & 0.25 & \textbf{0.03} & \textbf{0.03} & \textbf{0.33} & \textbf{0.03} &  4 \\
\bottomrule
Lens Inj.    & 0.93 &0.17 & 0.29 & \textbf{0.98} & 0.23 & 0.37 & \textbf{0.98} & \textbf{0.27} & \textbf{0.42} &  223 \\
\bottomrule
I/A          & 0.00 &0.00 & 0.00 & 0.00 & 0.00 & 0.00 & \textbf{0.50} & \textbf{0.29} & \textbf{0.37} &  19 \\
\bottomrule
Sec. Knife   & \textbf{1.00} &\textbf{0.06} & \textbf{0.11} & 0.79 & \textbf{0.06} & \textbf{0.11} & \textbf{1.00} & 0.01 & 0.02 &  187 \\
Micromanip.  & 0.03 &0.15 & 0.05 & \textbf{0.04} & \textbf{0.18} & \textbf{0.06} & 0.00 & 0.00 & 0.00 &  13 \\
Cap. Forceps & 0.00 &0.00 & 0.00 & 0.00 & 0.00 & 0.00 & 0.00 &0.00  & 0.00 &  20 \\
\hline
\hline
micro avg    & 0.96 &0.59 & 0.73 & 0.95 & 0.64 & 0.77 & 0.96 & 0.61  & 0.75   &  9457 \\
macro avg    & 0.67 &0.30 & 0.35 & 0.66 & 0.35 & 0.40 & 0.65 & 0.32  & 0.34   &  9457 \\
weighted avg & 0.97 &0.59 & 0.66 & 0.96 & 0.64 & 0.70 & 0.95 & 0.61  & 0.66   &  9457 \\

\bottomrule
\end{tabular}

}

\label{tab:my_label}
\end{table}

\clearpage
\begin{credits}
\subsubsection{\ackname} 
This work was conducted and fully financed by the Corporate Research and Technology department of Carl Zeiss AG.

\subsubsection{\discintname}
The authors have no competing interests to declare that are relevant to the content of this article.

\end{credits}

\clearpage

%
%
%
\bibliographystyle{splncs04}
\bibliography{ref}

\end{document}